\documentclass[letterpaper,10pt,journal,twoside]{IEEEtran}

\IEEEoverridecommandlockouts                              

\usepackage{balance}
\usepackage{color}

\usepackage{graphics} 
\usepackage{amsmath,amssymb,amsfonts} 
\usepackage{bm}
\usepackage{subcaption}	 
\usepackage{graphicx}
\usepackage{multirow}
\usepackage{array}

\usepackage{url}
\usepackage{hyperref}
\usepackage{xcolor}

\usepackage{float}
\usepackage{booktabs}
\usepackage{multirow}
\usepackage{textcomp}
\usepackage{setspace}
\usepackage{tabularx}
\usepackage{makecell}
\usepackage{colortbl}  
\usepackage{pifont}
\newcommand{\cmark}{\ding{51}}
\newcommand{\xmark}{\ding{55}}

\newcommand{\PAR}[1]{\vskip4pt \noindent{\bf #1~}}

\newcommand{\methodname}{VEOcc}

\definecolor{ceiling}{RGB}{214, 122, 122}
\definecolor{floor}{RGB}{122, 214, 122}
\definecolor{wall}{RGB}{122, 122, 214}
\definecolor{window}{RGB}{214, 214, 122}
\definecolor{chair}{RGB}{214, 122, 214}
\definecolor{bed}{RGB}{161, 222, 245}
\definecolor{sofa}{RGB}{201, 173, 217}
\definecolor{table}{RGB}{245, 184, 122}
\definecolor{tvs}{RGB}{153, 184, 122}
\definecolor{furniture}{RGB}{122, 184, 184}
\definecolor{objects}{RGB}{82, 138, 199}

\newif\ifshowrevision

\showrevisionfalse     

\ifshowrevision
    \newcommand{\rev}[1]{\textcolor{blue}{#1}}
\else
    \newcommand{\rev}[1]{#1}
\fi

\usepackage[backend=bibtex,natbib=true,style=numeric-comp,sorting=none,giveninits=true,maxbibnames=99,url=false,doi=false]{biblatex}
\title{VEOcc: Voxel-Centric Online Semantic Occupancy Prediction for Embodied Scene Understanding}

\author{
Ruoyu Wang$^{1,\dagger}$,
Yong Liu$^{1,\dagger,*}$,
Jiahan Li$^{1}$,
Sheng Tao,
Yuhang Lin$^{1}$,
Yukai Ma$^{1}$
\thanks{\rev{Manuscript received: May 28, 2026; Revised August 31, 2026; Accepted September 17, 2026.}}%
\thanks{\rev{This paper was recommended for publication by Editor Abhinav Valada upon evaluation of the Associate Editor and Reviewers' comments.}}%
\thanks{\rev{This work was supported by the National Natural Science Foundation of China (Grant No. 62525309)}}%
\thanks{\rev{$^{1}$ Institute of Cyber-Systems and Control, Zhejiang University, China.}}%
\thanks{\rev{$^{\dagger}$ These authors contributed equally to this work.}}%
\thanks{\rev{$^*$ Yong Liu is the corresponding author.}}%
\thanks{\rev{(Email: {\tt\small yongliu@iipc.zju.edu.cn}).}}%
\thanks{\rev{Digital Object Identifier (DOI): see top of this page.}}%
}
\begin{document}

\maketitle

\begin{abstract}
Crucial for autonomous exploration, online 3D occupancy prediction and mapping incrementally construct dense spatial representations on the fly. Embodied online occupancy prediction remains predominantly Gaussian-centric, despite the wide use of voxel representations for frame-wise scene completion. We present {\methodname}, to the best of our knowledge, the first voxel-centric framework for online embodied semantic occupancy prediction. It incrementally maintains a sparse global semantic voxel map from monocular observations and enables open-ended expansion without predefined scene bounds. To robustly integrate noisy multi-view predictions, we further introduce a Spatio-Temporal-Aware Online Update Strategy comprising Cross-Temporal Logit Aggregation (TLA) for short-term temporal consistency, Reliability-Aware Confidence Modulation (RCM) for spatial uncertainty calibration, and Confidence-Driven Incremental State Update (CSU) for robust global state assimilation. Extensive experiments on Occ-ScanNet and EmbodiedOcc-ScanNet demonstrate state-of-the-art performance among models of comparable scale in both local and embodied settings. Moreover, onboard deployment on a mobile robot validates practical online operation and long-horizon scalability, while results on self-collected handheld sequences demonstrate zero-shot generalization to unseen real-world environments.
Code and supplementary visualizations are available on our project page: \url{https://wryzju.github.io/VEOcc/}.
\end{abstract}




\section{Introduction}
\IEEEPARstart{O}{nline} 3D occupancy prediction and mapping are fundamental capabilities for embodied intelligence~\cite{11382039, xiaasurvey,roy2021machine, sun2024comprehensive, pfeifer2004embodied}, enabling robots to progressively build a coherent understanding of the global scene during exploration, thereby supporting high-level decision-making~\cite{huang2023voxposer,rana2023sayplan,gu2024conceptgraphs,bingi2025integrated,chen2025feature} in novel environments. Unlike traditional sparse geometric mapping~\cite{mur2015orb,campos2021orb,lai2022review,zhou2021improved}, dense 3D occupancy provides a unified representation of both geometry and categorical semantics. This comprehensive spatial awareness is essential for complex downstream tasks such as collision-free navigation and proactive object interaction.

Fundamentally, this task entails two tightly coupled phases: local prediction, which infers dense 3D structures from instantaneous ego-centric observations, and embodied prediction, which incrementally assimilates these partial views into a global map on the fly. From this perspective, an ideal system must overcome three core challenges: (1) ensuring accurate per-frame local prediction to guarantee global map fidelity; (2) enabling open-ended incremental construction without predefined spatial boundaries or dedicated mapping stages; and (3) robustly integrating long-horizon observations in the presence of noise and uncertainty.

\begin{figure}[t]
      \centering
      \includegraphics[width=\linewidth]{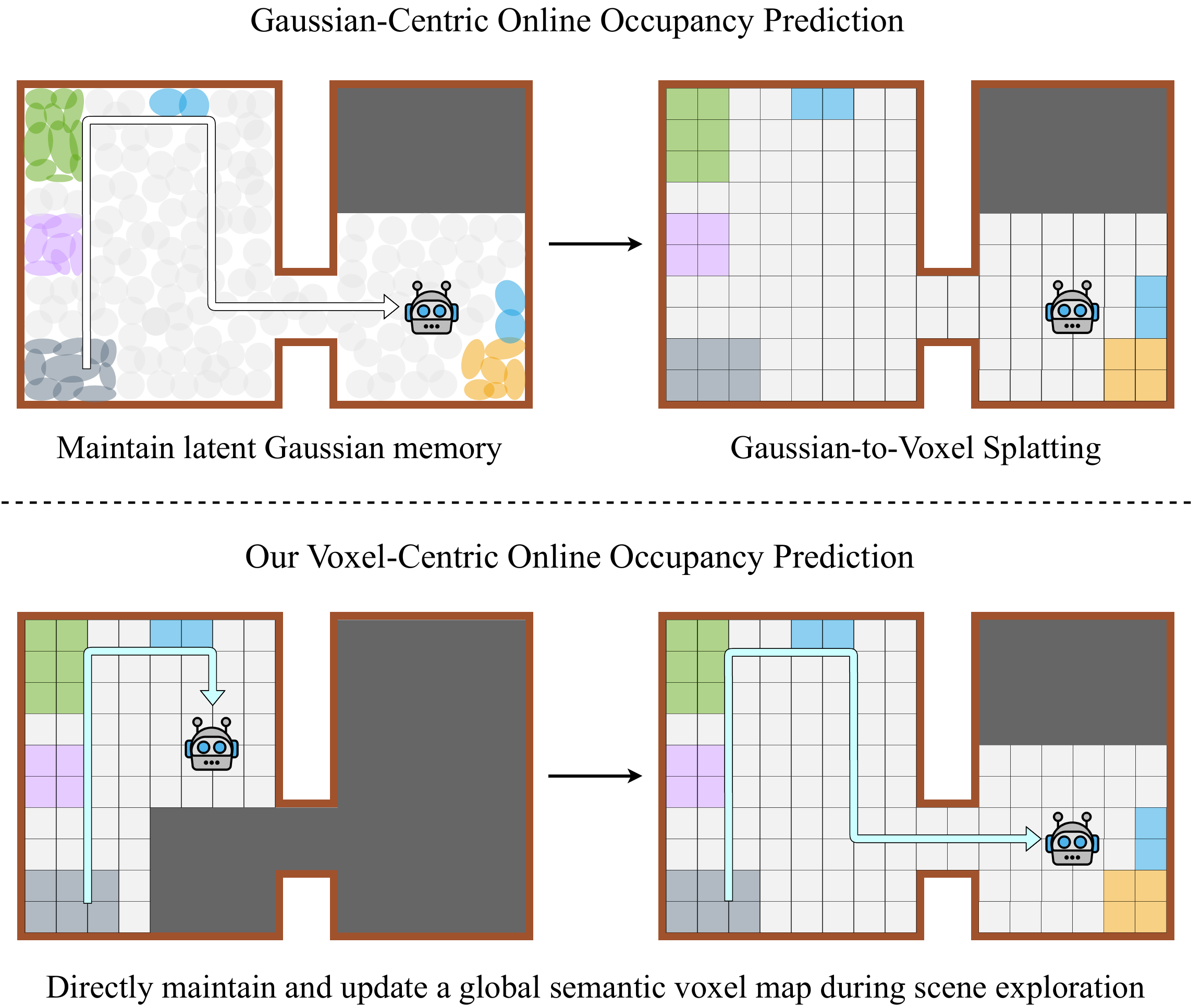}
      \captionsetup{font=small}
      \caption{\textbf{Comparison of Gaussian-centric and Voxel-centric online occupancy prediction.}
        Gaussian-centric methods maintain a latent Gaussian memory and render occupancy via Gaussian-to-voxel splatting, whereas VEOcc directly maintains and incrementally updates a global semantic voxel map during exploration.}
      \label{fig:teaser}
      \vspace{-15pt}
\end{figure}
Despite recent progress, approaches to embodied online semantic occupancy prediction have predominantly followed the Gaussian-centric paradigm~\cite{wu2025embodiedocc,wang2025embodiedocc++,zhang2025roboocc,qian2026tgsformer,Zhou_2026_CVPR}. Successive developments have improved prediction accuracy and, more recently, enabled scalable and borderless global mapping without predefined scene bounds. In contrast, voxel-centric representations remain largely unexplored for embodied online occupancy prediction, despite their regular spatial structure and strong geometric prediction capability in frame-wise scene completion. In particular, how to directly and reliably fuse sequential voxel predictions into a coherent, open-ended global occupancy map remains a research gap.

To fill this gap, we present \textbf{\methodname}, a voxel-centric framework that formulates online embodied occupancy prediction as a recursive perception-and-assimilation process. As illustrated in Fig.~\ref{fig:teaser}, {\methodname} departs from Gaussian-centric pipelines by directly maintaining semantic occupancy as a persistent global voxel state. At each timestep, a local occupancy predictor produces frame-wise voxel occupancy, which is then integrated through a Spatio-Temporal-Aware Online Update Strategy. Cross-Temporal Logit Aggregation (TLA) fuses short-term predictions, Reliability-Aware Confidence Modulation (RCM) calibrates spatially varying observation reliability, and Confidence-Driven Incremental State Update (CSU) recursively updates the long-term map. Together, these modules suppress cross-view inconsistencies and enable coherent online semantic mapping.

Extensive experiments validate our design across both local and embodied settings. On Occ-ScanNet~\cite{yu2024monocular} and EmbodiedOcc-ScanNet~\cite{wu2025embodiedocc}, VEOcc outperforms recent Gaussian-centric baselines, including TGSFormer~\cite{qian2026tgsformer} and GPOcc-DPT~\cite{Zhou_2026_CVPR}, achieving state-of-the-art performance among models of comparable scale. Crucially, onboard deployment on a mobile robot validates practical online operation and long-horizon scalability, while zero-shot results on self-collected sequences demonstrate generalization to unseen real-world environments.

Our contributions are summarized as follows:
\begin{itemize}
    \item We propose {\methodname}, to the best of our knowledge, the first voxel-centric framework designed for online embodied semantic occupancy prediction, bridging frame-wise voxel prediction and persistent global mapping while supporting open-ended incremental expansion.
    
    \item We design a Spatio-Temporal-Aware Online Update Strategy for reliable voxel-state assimilation. It combines Cross-Temporal Logit Aggregation (TLA) for short-term fusion, Reliability-Aware Confidence Modulation (RCM) for spatial observation reliability calibration, and Confidence-Driven Incremental State Update (CSU) for long-term global state assimilation.
    
    \item Extensive experiments on Occ-ScanNet~\cite{yu2024monocular} and EmbodiedOcc-ScanNet~\cite{wu2025embodiedocc} demonstrate state-of-the-art performance among models of comparable scale in both local and embodied settings. Onboard robotic deployment and zero-shot evaluations on self-collected handheld sequences further validate practical online operation and generalization to unseen real-world environments.
\end{itemize}

%
%
%
%
%
%
%

\section{RELATED WORK}
\label{sec:related work}\subsection{3D occupancy prediction.}
3D occupancy prediction jointly infers scene geometry and semantics from sensor observations. 
\rev{Existing methods mainly fall into two categories~\cite{deng2026best}: dense representations~\cite{cao2022monoscene,zhang2023occformer,yu2024context,wang2025l2cocc, li2025voxdet, li2023voxformer,jiang2023symphonize,wei2023surroundocc, huang2023tri,chen2026vggtocc}} and sparse representations~\cite{huang2024gaussianformer,shi2025odg, boeder2025gaussianflowocc, qian2026splatssc}. 
Dense approaches typically model the scene with BEV (bird’s-eye view) or voxel grids, where the view transformer plays a central role. Some methods~\cite{zhang2023occformer,yu2024context,wang2025l2cocc,li2025voxdet} follow the LSS (Lift-Splat-Shoot~\cite{philion2020lift}) paradigm and lift image features into 3D via depth distributions, while others~\cite{li2023voxformer,jiang2023symphonize,huang2023tri, wei2023surroundocc} directly learn dense scene representations with transformer-based architectures. However, these methods often incur substantial computational cost when scaling to large scenes. 
In parallel, sparse representations have gained increasing attention. GaussianFormer~\cite{huang2024gaussianformer} first introduces 3D Gaussian primitives into occupancy prediction, and subsequent works such as GaussianFlowOcc~\cite{boeder2025gaussianflowocc} leverage 2D semantic and depth supervision to guide Gaussian learning. ODG~\cite{shi2025odg} further adopts separate branches for dynamic and static categories, optimized with both 2D and 3D losses. 
The main advantage of 3D Gaussian representations lies in their efficiency and compatibility with 2D supervision from depth and semantic maps.
Nevertheless, they do not consistently demonstrate clear advantages in accuracy over contemporary voxel-centric methods.

\subsection{Embodied online occupancy prediction.} EmbodiedOcc~\cite{wu2025embodiedocc} first formulated the embodied online occupancy prediction task and adopted 3D Gaussian as the scene representation. It initializes a global scene with random Gaussians according to the scene extent and updates the representation incrementally as new observations arrive. Building upon this line of work, EmbodiedOcc++~\cite{wang2025embodiedocc++} proposed adaptive geometric constraints to better align Gaussians with planar and curved surfaces, while RoboOcc~\cite{zhang2025roboocc} introduced an opacity-guided autoencoder (OSE) to alleviate semantic ambiguity caused by overlapping Gaussians, leading to improved performance. 
More recently, TGSFormer~\cite{qian2026tgsformer} and GPOcc~\cite{Zhou_2026_CVPR} extended Gaussian-centric approaches to scalable and borderless global mapping without predefined scene bounds. EmbodiedOcc also reported local prediction results from several voxel-centric methods~\cite{cao2022monoscene,yu2024monocular,wei2023surroundocc,huang2023tri}.
While their performance remains limited, research on voxel-centric dense representations for online occupancy prediction is still relatively underexplored. Our work aims to fill this gap and demonstrates the advantages of voxel representations in indoor scenarios.

\section{Methodology}
\label{sec:methodology}
\begin{figure*}[t]
      \centering
      \includegraphics[width=\linewidth]{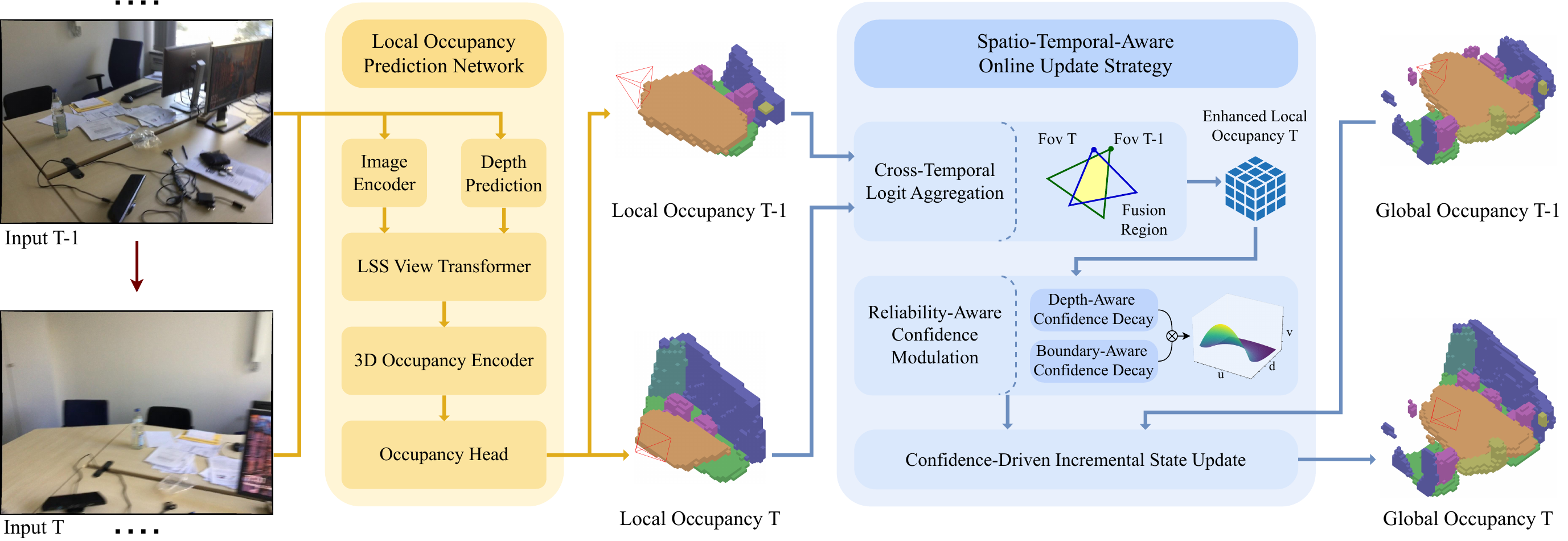}
      \captionsetup{font={small}}
      \caption{\textbf{The overall framework of our proposed {\methodname}.} Given a sequence of monocular images, a voxel-centric network first predicts frame-wise local occupancy, which is then incrementally assimilated into a global occupancy grid via the proposed Spatio-Temporal-Aware Online Update Strategy. Within this strategy, Cross-Temporal Logit Aggregation (TLA) enforces short-term temporal consistency, Reliability-Aware Confidence Modulation (RCM) calibrates spatial uncertainties, and Confidence-Driven Incremental State Update (CSU) recursively integrates these refined observations to ensure a robust global semantic state.
      }
      \vspace{-10pt}
      \label{fig:overview}
\end{figure*}

\subsection{Problem Formulation}
\label{problem_formualtion}
\textbf{Local Occupancy Prediction} predicts the semantic occupancy within the current field of view, denoted as $\boldsymbol{Y}_{t}^{\text{local}} \in \mathbb{R}^{X_l \times Y_l \times Z_l \times N_c}$. 
\textbf{Embodied Occupancy Prediction} constructs a global semantic occupancy map $\boldsymbol{Y}_t \in \mathbb{R}^{X_g \times Y_g \times Z_g \times N_c}$ by integrating sequential observations. $N_c$ indicates the number of predefined semantic categories.
At timestep $t$, given an image $\boldsymbol{I}_t \in \mathbb{R}^{H \times W \times 3}$ and geometric priors 
$\boldsymbol{G}_t = \{\boldsymbol{K}_t, \boldsymbol{T}_t, \boldsymbol{o}_t, \boldsymbol{s}, \boldsymbol{v}\}$, the tasks are formulated as:
\begin{equation}
\boldsymbol{Y}_{t}^{\text{local}} = \mathcal{F}_{\text{local}}(\boldsymbol{I}_t, \boldsymbol{G}_t), \quad
\boldsymbol{Y}_t = \mathcal{F}_{\text{embodied}}(\mathcal{O}_{t-1}, \boldsymbol{I}_t, \boldsymbol{G}_t).
\end{equation}

Here, $\boldsymbol{K}_t$ and $\boldsymbol{T}_t$ denote the camera intrinsics and pose. 
$\boldsymbol{o}_t$ is the origin of the local volume in the global coordinate system, which is updated according to the camera pose at each timestep. 
$\boldsymbol{s} \in \mathbb{R}^3$ defines the physical extent of the local volume in meters, discretized into $X_l \times Y_l \times Z_l$ with voxel size $\boldsymbol{v} \in \mathbb{R}^3$.
$X_g, Y_g, Z_g$ denote the resolution of the global occupancy grid, whose spatial extent is determined online as the scene is explored.
$\mathcal{O}_{t-1}$ indicates the underlying global occupancy representation maintained by the system (detailed in Section~\ref{embodied_pred}). 
This formulation models embodied occupancy prediction as a sequential state update process under partial observations.

\subsection{Overview of {\methodname}}
As illustrated in Fig.~\ref{fig:overview}, VEOcc processes a sequence of monocular images to perform online voxel-centric occupancy prediction and mapping. The framework first estimates frame-wise local occupancy via a voxel-centric occupancy prediction network (Section~\ref{local_pred}), and then incrementally maintains a global occupancy grid through a Spatio-Temporal-Aware Online Update Strategy (Section~\ref{embodied_pred}). 
This strategy synergistically integrates three modules—\textbf{TLA, RCM, and CSU}—to recursively assimilate noisy multi-view observations into a robust global semantic state.

\subsection{Local Occupancy Prediction Network}
\label{local_pred}
Given a monocular image at timestep $t$, we first extract multi-scale 2D features using an EfficientNet-B7~\cite{tan2019efficientnet} backbone with a Feature Pyramid Network (FPN)~\cite{lin2017feature} neck, producing a unified feature map $\boldsymbol{F}_t \in \mathbb{R}^{H \times W \times C}$. 
In parallel, a depth foundation model (Depth Anything V2~\cite{yang2024depth}) predicts a dense depth map $\boldsymbol{D}_t \in \mathbb{R}^{H \times W}$. 
We then adopt a standard voxel-centric pipeline for local occupancy prediction.

\PAR{2D-to-3D Lifting.}
To handle diverse unconstrained viewpoints in embodied exploration, we adopt an LSS~\cite{philion2020lift} view transformer that explicitly incorporates camera intrinsics and extrinsics, reducing the learning burden of implicit geometric reasoning.
Following the design of L2COcc~\cite{wang2025l2cocc}, we employ a Depth Net and a Context Net to predict a discrete depth distribution $\boldsymbol{D}_t^\text{dist} \in \mathbb{R}^{H \times W \times D}$ and context features $\boldsymbol{C}_t \in \mathbb{R}^{H \times W \times C}$, respectively, conditioned on the predicted depth $\boldsymbol{D}_t$. 
The two are then combined via an outer product to form the LSS feature volume $\boldsymbol{V}_t = \boldsymbol{D}_t^\text{dist} \otimes \boldsymbol{C_t}$.
Next, we compute the 3D coordinates of each local voxel based on the local origin $\boldsymbol{o}_t$, physical extent $\boldsymbol{s}$, and voxel size $\boldsymbol{v}$. 
The LSS feature volume $\boldsymbol{V}$ is then projected onto the corresponding 3D grid using camera intrinsics $\boldsymbol{K}_t$ and pose $\boldsymbol{T}_t$, producing the 3D feature representation
\begin{equation}
\boldsymbol{F}^{\text{3D}}_t \in \mathbb{R}^{X_l \times Y_l \times Z_l} = \text{Splat}(\boldsymbol{V}_t, \boldsymbol{G}_t).
\end{equation}

\PAR{Occupancy Prediction.}
The lifted voxel features are further encoded by a 3D ResNet~\cite{he2016deep} backbone and a 3D FPN neck to aggregate context across scales. 
A lightweight 3D occupancy head then predicts per-voxel semantic logits $\boldsymbol{Z}_t^\text{local} \in \mathbb{R}^{X_l \times Y_l \times Z_l \times N_c}$ for all occupancy classes.
The final semantic prediction $Y_t^\text{local}$ is obtained by applying a softmax over the class dimension.
We optimize local prediction with a composite objective aligned with prior works~\cite{wu2025embodiedocc,zhang2025roboocc}:
\begin{equation}
\mathcal{L}_{\text{local}} = \lambda_{\text{focal}}\mathcal{L}_{\text{focal}} + \mathcal{L}_{\text{lovasz}} + \mathcal{L}^{\text{sem}}_{\text{scal}} + \mathcal{L}^{\text{geo}}_{\text{scal}},
\end{equation}
where $\mathcal{L}_{\text{focal}}$ (with weight $\lambda_{\text{focal}}=100$), $\mathcal{L}_{\text{lovasz}}$, $\mathcal{L}_{\text{scal}}^{\text{sem}}$, and $\mathcal{L}_{\text{scal}}^{\text{geo}}$ correspond to the standard focal, Lovasz-Softmax, and scene-class affinity losses, respectively. 

\subsection{Spatio-Temporal-Aware Online Update Strategy}
\label{embodied_pred}

\PAR{Global Occupancy Representation.}
Instead of maintaining a dense global tensor throughout exploration, we represent global occupancy as a sparse set of tuples
\begin{equation}
\mathcal{O}_t = \{(\boldsymbol{p}_i, \boldsymbol{s}_{i,t}, c_{i,t}, n_{i,t})\}_{i=1}^{N_t},
\end{equation}
where $\boldsymbol{p}_i \in \mathbb{R}^3$ denotes the voxel coordinate, $\boldsymbol{s}_{i,t} \in \mathbb{R}^{N_c}$ represents the semantic probability, $c_{i,t} \in \mathbb{R}$ is the confidence score, and $n_{i,t} \in \mathbb{N}$ indicates the number of observations of the voxel.
This sparse formulation ensures memory-efficient map expansion, allowing $\mathcal{O}_t$ to be flexibly converted into a dense grid $\boldsymbol{Y}_t$ strictly on-demand for downstream tasks.
The grid resolution $(X_g, Y_g, Z_g)$ is determined by the spatial extent of $\{\boldsymbol{p}_i\}_{i=1}^{N_t}$ along each axis, based on which an empty grid $\boldsymbol{Y}_t$ is initialized. Each $\boldsymbol{p}_i$ is then mapped to its corresponding discrete voxel, where the semantic label is assigned as
\begin{equation}
\boldsymbol{Y}_t(\mathbf{p}_i) = \arg\max \mathbf{s}_{i,t}.
\end{equation}

\PAR{Cross-Temporal Logit Aggregation}
At timestep $t$, we first identify the set of visible voxels based on geometric priors $\boldsymbol{G}_t$, treating their center coordinates in the global coordinate system as candidates $\boldsymbol{P}_t=\{\boldsymbol{p}_i\}_{i=1}^{n_t}$ for the global update, where $n_t$ denotes the number of visible voxels. To address the inherent spatial misalignment between the discrete global grid and the moving ego-centric frame, each $\boldsymbol{p}_i$ is projected into the continuous local space. The corresponding logit and feature representations $\boldsymbol{Z}_t=\{\boldsymbol{z}_{i,t}\}_{i=1}^{n_t}$ and $\boldsymbol{F}_t^\text{voxel}=\{\boldsymbol{f}_{i,t}^{\text{voxel}}\}_{i=1}^{n_t}$ are then sampled via trilinear interpolation over the local volumes
$\boldsymbol{Z}_t^\text{local}$ and $\boldsymbol{F}_t^{3D}$, ensuring precise spatial alignment prior to temporal fusion.
We further define a temporally consistent subset $\boldsymbol{P}_t^{\text{temp}}=\{\boldsymbol{p}_i\}_{i=1}^{n_\text{temp}} \subseteq \boldsymbol{P}_t$, where $n_{\text{temp}}$ denotes the number of voxels also visible at timestep $t-1$. The corresponding representations $\boldsymbol{Z}_{t-1}=\{\boldsymbol{z}_{i,t-1}\}_{i=1}^{n_\text{temp}}$ and $\boldsymbol{F}_{t-1}^{\text{voxel}}=\{\boldsymbol{f}_{i,t-1}^{\text{voxel}}\}_{i=1}^{n_\text{temp}}$ are obtained from the previous local volumes.
\begin{figure}[t]
      \centering
      \includegraphics[width=\linewidth]{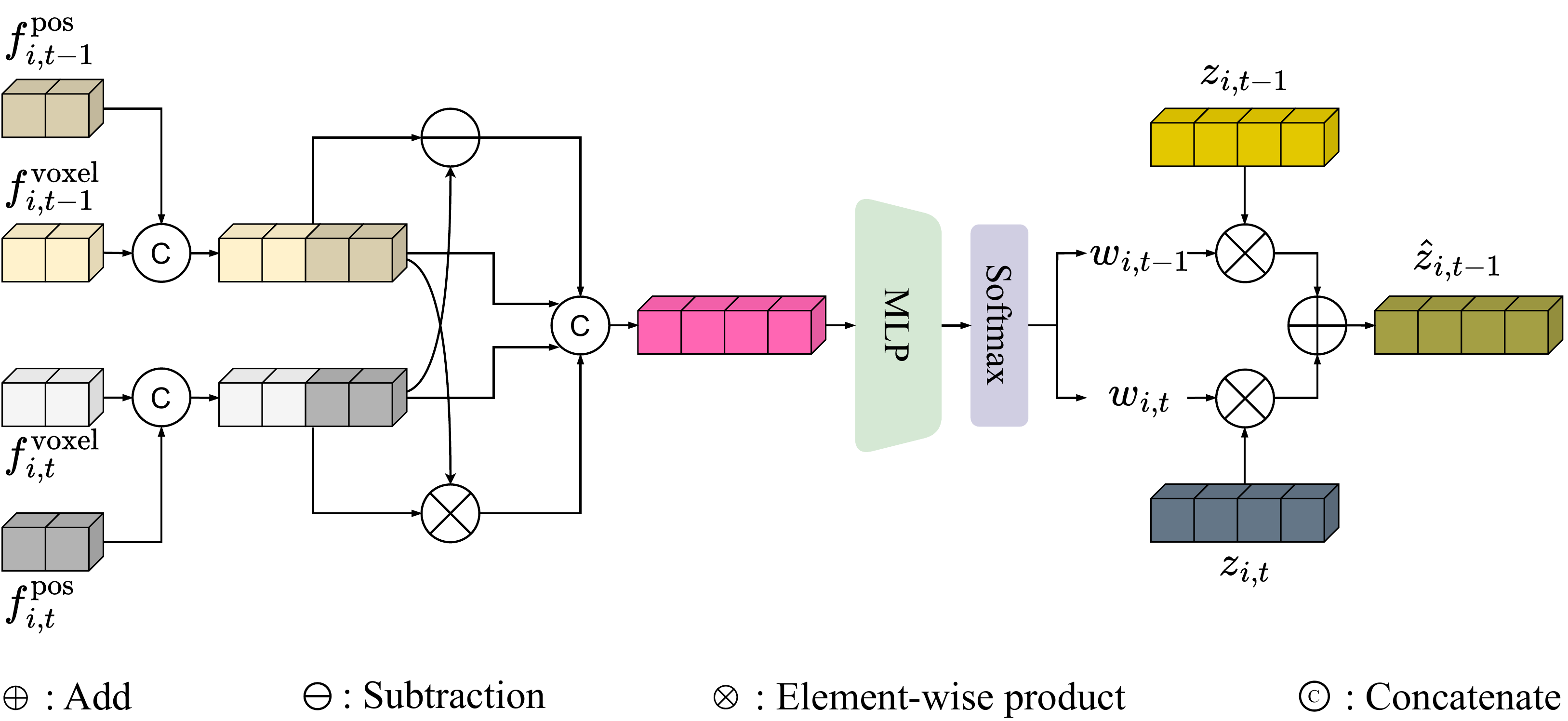}
      \captionsetup{font=small}
      \caption{\textbf{Design of Cross-Temporal Logit Aggregation (TLA).} This module adaptively aggregates semantic logits from adjacent frames by explicitly modeling cross-view discrepancies in both feature representations and spatial contexts.}
      \vspace{-12pt}
      \label{fig:tlf}
\end{figure}
Considering occlusion and depth ambiguity, the reliability of predictions for the same voxel may vary across viewpoints. We thus perform temporal fusion on $\boldsymbol{p}_i \in \boldsymbol{P}_t^{\text{temp}}$ by jointly leveraging their current and previous representations for logit aggregation. As illustrated in Fig.~\ref{fig:tlf}, we thus construct temporal pairwise features by combining current and previous representations, together with their difference and interaction terms. To incorporate geometric context, each $\boldsymbol{p}_i$ is projected into the camera and image spaces at both timesteps, followed by sinusoidal positional encoding to produce $\boldsymbol{f}^\text{pos}_{i,t}$ and $\boldsymbol{f}^\text{pos}_{i,t-1}$.
Based on the discrepancies in both feature representations and spatial cues, a lightweight MLP predicts normalized fusion weights via softmax, which reflect the relative reliability of adjacent observations and are used to adaptively fuse the temporal representations:
\begin{equation}
[w_{i,t}, w_{i,t-1}] = \mathrm{Softmax}(\phi(\boldsymbol{f}_{i,t}^{\text{voxel}},\boldsymbol{f}_{i,t}^{\text{pos}},\boldsymbol{f}_{i,t-1}^{\text{voxel}},\boldsymbol{f}_{i,t-1}^{\text{pos}})).
\end{equation}
The fused semantic logits $\hat{\boldsymbol{Z}}_t=\{\hat{\boldsymbol{z}}_{i,t}\}_{i=1}^{n_\text{temp}}$ are obtained via weighted aggregation:
\begin{equation}
\hat{\boldsymbol{z}}_{i,t} = w_{i,t} \boldsymbol{z}_{i,t} + w_{i,t-1} \boldsymbol{z}_{i,t-1}.
\end{equation}
Finally, the corresponding logits in $\boldsymbol{Z}_t$ are replaced with $\hat{\boldsymbol{Z}}_t$ as enhanced predictions for global update.

During training, fused voxel logits are supervised using a cross-entropy (CE) loss computed on the corresponding semantic labels $\{\boldsymbol{y}_i^\text{gt}\}_{i=1}^{n_\text{temp}}$:
\begin{equation}
\mathcal{L}_\text{embodied} = \frac{1}{n_\text{temp}} \sum_{i=1}^{n_\text{temp}} \text{CE}(\hat{\boldsymbol{z}}_{i,t}, \boldsymbol{y}_i^{\text{gt}}).
\end{equation}

\PAR{Reliability-Aware Confidence Modulation.}
The accuracy of local occupancy prediction is highly correlated with both semantic feature quality and depth estimation reliability. However, in monocular settings, feature and geometric uncertainties are not uniformly distributed across the image. Specifically, voxel predictions near image boundaries tend to suffer from weaker semantic cues due to incomplete contextual information, while voxels at large depth often exhibit degraded geometric accuracy caused by increased depth ambiguity and projection instability.

To explicitly model such variations, we define a confidence set $\boldsymbol{C}_t = \{\tilde{c}_{i,t}\}_{i=1}^{n_t}$ corresponding to the coordinate candidates $\boldsymbol{P}_t$. Each $c_i$ reflects the reliability of coordinate $\boldsymbol{p}_i$ in local occupancy prediction and is designed to decrease for distant and boundary-affected regions. Confidence values are computed via Depth- and Boundary-Aware Confidence Decay from depth information and image-space geometry:
\begin{equation}
\tilde{c}_{i,t} = \mathrm{clip}\left(\exp(-\alpha d_{i,t})\cdot \exp(-\beta b_{i,t}),\ c_{\min},\ 1\right),
\end{equation}
\rev{where $d_{i,t}$ denotes the depth of voxel $p_i$ obtained from its projection onto the camera ray, and $b_{i,t}=1-d^{\mathrm{edge}}_{i,t}/r$ denotes the normalized boundary proximity, where $d^{\mathrm{edge}}_{i,t}=\min\{u_{i,t},W-1-u_{i,t},v_{i,t},H-1-v_{i,t}\}$ and $r=\max\{\min(W,H)/2,1\}$.
The coefficients $\alpha=0.1$ and $\beta=1.5$ control the sensitivity of confidence decay with respect to geometric and boundary-related uncertainty. Voxels with extreme confidence suppression are handled via the lower bound $c_{\min}=0.01$. }


\begin{figure}[t]
      \centering
      \includegraphics[width=\linewidth]{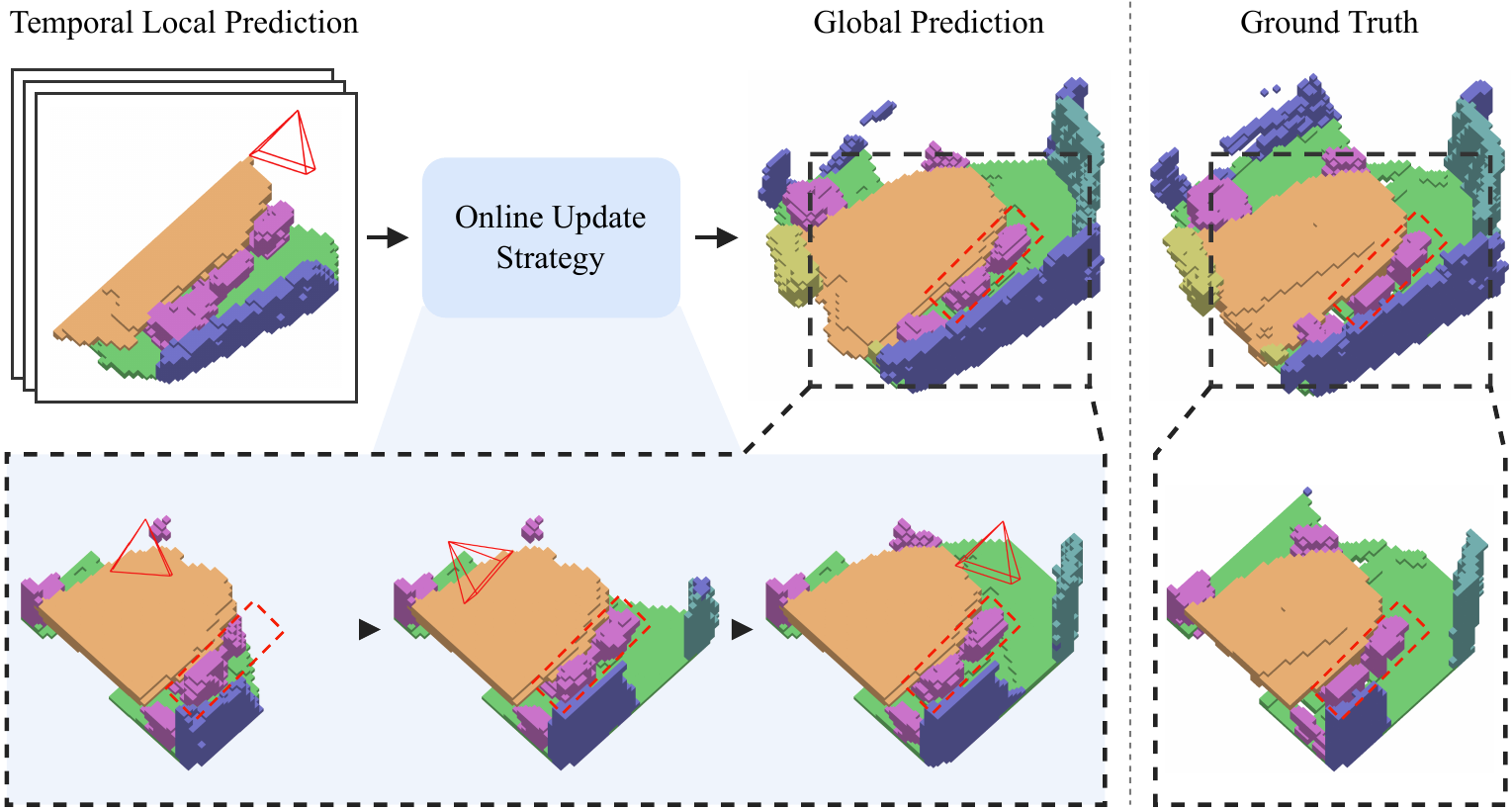}
      \captionsetup{font=small}
      \caption{\textbf{Visualization of the proposed Spatio-Temporal-Aware Online Update Strategy.} As the global occupancy is incrementally built over time, unreliable predictions from early observations are progressively refined by incoming reliable observations, demonstrating the efficacy of the proposed strategy.}
      \label{fig:GFS}
      \vspace{-12pt}
\end{figure}

\begin{table*}
    \captionsetup{font={small}}
	\setlength{\tabcolsep}{0.0035\linewidth}
	\newcommand{\classfreq}[1]{{~\tiny(\semkitfreq{#1}\%)}}  %
	\centering
    \caption{\textbf{Local Prediction Performance} on the Occ-ScanNet dataset. We mark the best score in \textbf{bold}.
    }
   \resizebox{1\linewidth}{!}{
	\begin{tabular}{l |c |c c |c c c c c c c c c c c}
 
		\toprule
		Method
        & \makecell[c]{Dataset}
		& \makecell[c]{IoU}
        & \makecell[c]{mIoU}
		& \rotatebox{90}{\textcolor{ceiling}{$\blacksquare$} ceiling}
		& \rotatebox{90}{\textcolor{floor}{$\blacksquare$} floor}
		& \rotatebox{90}{\textcolor{wall}{$\blacksquare$} wall} 
		& \rotatebox{90}{\textcolor{window}{$\blacksquare$} window} 
		& \rotatebox{90}{\textcolor{chair}{$\blacksquare$} chair} 
		& \rotatebox{90}{\textcolor{bed}{$\blacksquare$} bed} 
		& \rotatebox{90}{\textcolor{sofa}{$\blacksquare$} sofa} 
		& \rotatebox{90}{\textcolor{table}{$\blacksquare$} table} 
        & \rotatebox{90}{\textcolor{tvs}{$\blacksquare$} tvs} 
        & \rotatebox{90}{\textcolor{furniture}{$\blacksquare$} furniture} 
		& \rotatebox{90}{\textcolor{objects}{$\blacksquare$} objects} \\

		\midrule
        
        
        MonoScene~\cite{cao2022monoscene} & \multirow{8}{*}{Occ-ScanNet} & 41.60 & 24.62 & 15.17 & 44.71 & 22.41 & 12.55 & 26.11 & 27.03 & 35.91 & 28.32 & 6.57 & 32.16 & 19.84 \\
        EmbodiedOcc~\cite{wu2025embodiedocc} & & 53.95 & 45.48 & 40.90 & 50.80 & 41.90 & 33.00 & 41.20 & 55.20 & 61.90 & 43.80 & 35.40 & 53.50 & 42.90 \\
        EmbodiedOcc++~\cite{wang2025embodiedocc++} & & 54.90 & 46.20 & 36.40 & 53.10 & 41.80 & 34.40 & 42.90 & 57.30 & 64.10 & 45.20 & 34.80 & 54.20 & 44.10 \\
        RoboOcc~\cite{zhang2025roboocc} & & 56.48 & 47.67 & 45.36 & 53.49 & 44.35 & 34.81 & 43.38 & 56.93 & 63.35 & 46.35 & 36.12 & 55.48 & 44.78 \\
        SplatSSC~\cite{qian2026splatssc} & & 62.83 & 51.83 & 49.10 & 59.00 & 48.30 & 38.80 & 47.40 & 62.40 & 67.00 & 49.50 & 42.60 & 60.70 & 45.40  \\
        TGSFormer~\cite{qian2026tgsformer} & & 64.42 & 54.73 & \textbf{54.20} & 61.30 & \textbf{50.50} & 41.30 & 49.50 & 64.50 & 69.30 & 52.40 & 47.10 & \textbf{63.00} & 48.80\\
        GPOcc-DPT~\cite{Zhou_2026_CVPR} & & 56.96 & 51.88 & 51.42 & 50.35 & 46.97 & 41.84 & 46.98 & 60.39 & 66.16 & 50.51 & 47.97 & 58.88 & 49.23 \\
        {\methodname} (Ours) & & \textbf{64.55} & \textbf{55.49} & 53.20 & \textbf{64.00} & 49.00 & \textbf{42.90} & \textbf{50.80} & \textbf{65.50} & \textbf{70.50} & \textbf{53.70} & \textbf{48.30} & 62.60 & \textbf{50.00}  \\
        \midrule
        \bottomrule
	\end{tabular}} \\
	\label{table:local_pred_main}
\end{table*}

\begin{table*}
    \captionsetup{font={small}}
	\setlength{\tabcolsep}{0.0035\linewidth}
	\newcommand{\classfreq}[1]{{~\tiny(\semkitfreq{#1}\%)}}  %
	\centering
    \caption{\textbf{Embodied Prediction Performance} on the EmbodiedOcc-ScanNet dataset. We mark the best score in \textbf{bold}.
    }
   \resizebox{1\linewidth}{!}{
	\begin{tabular}{l |c |c c |c c c c c c c c c c c}
 
		\toprule
		Method
        & \makecell[c]{Dataset}
		& \makecell[c]{IoU}
        & \makecell[c]{mIoU}
		& \rotatebox{90}{\textcolor{ceiling}{$\blacksquare$} ceiling}
		& \rotatebox{90}{\textcolor{floor}{$\blacksquare$} floor}
		& \rotatebox{90}{\textcolor{wall}{$\blacksquare$} wall} 
		& \rotatebox{90}{\textcolor{window}{$\blacksquare$} window} 
		& \rotatebox{90}{\textcolor{chair}{$\blacksquare$} chair} 
		& \rotatebox{90}{\textcolor{bed}{$\blacksquare$} bed} 
		& \rotatebox{90}{\textcolor{sofa}{$\blacksquare$} sofa} 
		& \rotatebox{90}{\textcolor{table}{$\blacksquare$} table} 
        & \rotatebox{90}{\textcolor{tvs}{$\blacksquare$} tvs} 
        & \rotatebox{90}{\textcolor{furniture}{$\blacksquare$} furniture} 
		& \rotatebox{90}{\textcolor{objects}{$\blacksquare$} objects} \\
        
        \midrule
        
        EmbodiedOcc~\cite{wu2025embodiedocc}
        & \multirow{6}{*}{\makecell{EmbodiedOcc- \\ ScanNet}}
        & 51.52 & 42.53 & 22.70 & 44.60 & 37.40 & 38.00 & 50.10 & 56.70 & 59.70 & 35.40 & 38.40 & 52.00 & 32.90 \\
        
        EmbodiedOcc++~\cite{wang2025embodiedocc++} &
        & 52.20 & 43.60 & 27.90 & 43.90 & 38.70 & 40.60 & 49.00 & 57.90 & 59.20 & 36.80 & 37.80 & 53.50 & 34.10 \\
        
        RoboOcc~\cite{zhang2025roboocc} &
        & 53.30 & 44.05 & 21.94 & 44.57 & 39.54 & 38.48 & 51.28 & 57.04 & 63.09 & 36.70 & 43.05 & 54.42 & 34.38 \\
        
        TGSFormer~\cite{qian2026tgsformer} &
        & 54.42 & 45.29 & 31.00 & 39.90 & 39.90 & 32.70 & 52.00 & 66.10 & 53.00 & 47.30 & 38.70 & 59.60 & 37.90 \\
        
        GPOcc-DPT~\cite{Zhou_2026_CVPR} &
        & 56.39 & 51.22 & \textbf{40.80} & 48.78 & \textbf{45.62} & 43.26 & 50.08 & 63.97 & 67.72 & 48.36 & \textbf{48.77} & 60.46 & 45.63 \\
        
        \rev{{\methodname} (Ours)} &
        & \textbf{62.21} & \textbf{53.00}
        & 29.90 & \textbf{62.60} & 45.60 & \textbf{44.60}
        & \textbf{54.10} & \textbf{68.10} & \textbf{69.70}
        & \textbf{54.30} & 45.90 & \textbf{61.60} & \textbf{46.40} \\
        
        \bottomrule
	\end{tabular}} \\
	\label{table:embodied_pred_main}
    \vspace{-10pt}
\end{table*}

\PAR{Confidence-Driven Incremental State Update.}
While local temporal aggregation (TLA) operates effectively on logits due to the proximity of adjacent feature spaces, long-term global fusion is highly susceptible to logit variance drift over time. To ensure numerical stability across distant viewpoints, we first convert the enhanced voxel logits $\boldsymbol{Z}_t$ into normalized semantic probabilities via softmax before global integration:
\begin{equation}
\tilde{\boldsymbol{S}}_t=\{\tilde{\boldsymbol{s}}_{i,t}\}_{i=1}^{n_t},
\qquad
\tilde{\boldsymbol{s}}_{i,t}=\mathrm{Softmax}(\boldsymbol{z}_{i,t}),
\end{equation}
where $\tilde{\boldsymbol{s}}_{i,t}\in\mathbb{R}^{N_c}$ denotes the semantic probability associated with voxel coordinate $\boldsymbol{p}_i$.
The semantic probabilities are then incrementally integrated into the global occupancy representation $\mathcal{O}_{t-1}$. 

For a voxel coordinate $\boldsymbol{p}_i$ that is observed for the first time, we directly initialize its global state as
\begin{equation}
(\boldsymbol{p}_i,\boldsymbol{s}_{i,t},c_{i,t},n_{i,t})= (\boldsymbol{p}_i,\tilde{\boldsymbol{s}}_{i,t},\tilde{c}_{i,t},1).
\end{equation}
Otherwise, if $\boldsymbol{p}_i$ already exists in $\mathcal{O}_{t-1}$, we perform confidence-aware semantic fusion between the historical observation and the current prediction:
\begin{equation}
\boldsymbol{s}_{i,t} = \frac{\lambda n_{i,t-1}c_{i,t-1}\boldsymbol{s}_{i,t-1} + \tilde{c}_{i,t}\boldsymbol{\tilde{s}}_{i,t}}{\lambda n_{i,t-1}c_{i,t-1}+\tilde{c}_{i,t}},
\end{equation}
where $\lambda$ is a balancing coefficient controlling the reliance on historical observations.
The confidence score and observation count are updated as
\begin{equation}
c_{i,t} = \frac{\lambda n_{i,t-1}c_{i,t-1}+\tilde{c}_{i,t}}{\lambda n_{i,t-1}+1}, \quad n_{i,t} = n_{i,t-1}+1
\end{equation}

In this manner, reliable observations contribute more strongly to the global occupancy representation, while noisy predictions from distant or boundary regions are progressively suppressed during long-term embodied exploration (See Fig.~\ref{fig:GFS}). While we set $\lambda = 1$ to strictly accumulate observations for static benchmarks, adopting a smaller $\lambda$ effectively controls historical inertia, gracefully decaying outdated priors to correct early noisy predictions.

\section{Experiments}
\label{sec:experiement}

\begin{figure*}[t]
      \centering
      \includegraphics[width=\linewidth]{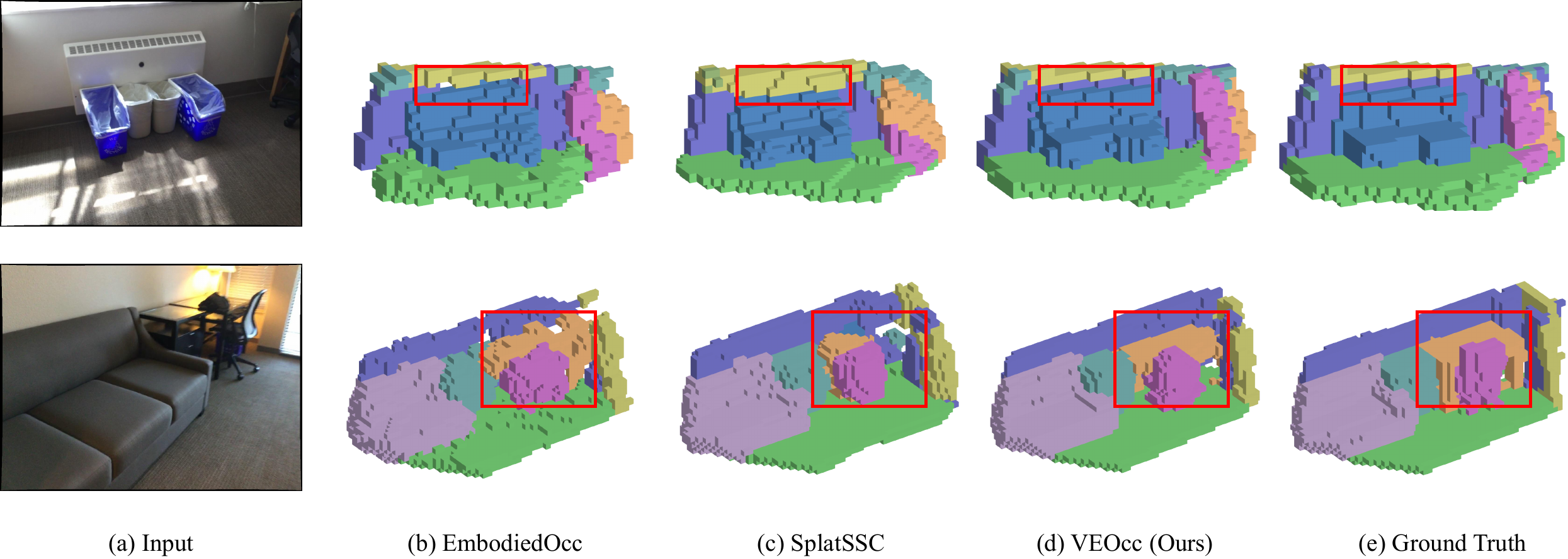}
      \captionsetup{font={small}}
      \caption{\textbf{Qualitative results of local occupancy prediction on Occ-ScanNet.} Our VEOcc achieves noticeably better prediction quality in object details, structural boundaries, occluded regions, and overall spatial smoothness compared with previous Gaussian-centric approaches.}
      \label{fig:vis_local_pred}
      \vspace{-10pt}
\end{figure*}
\begin{figure*}[t]
      \centering
      \includegraphics[width=\linewidth]{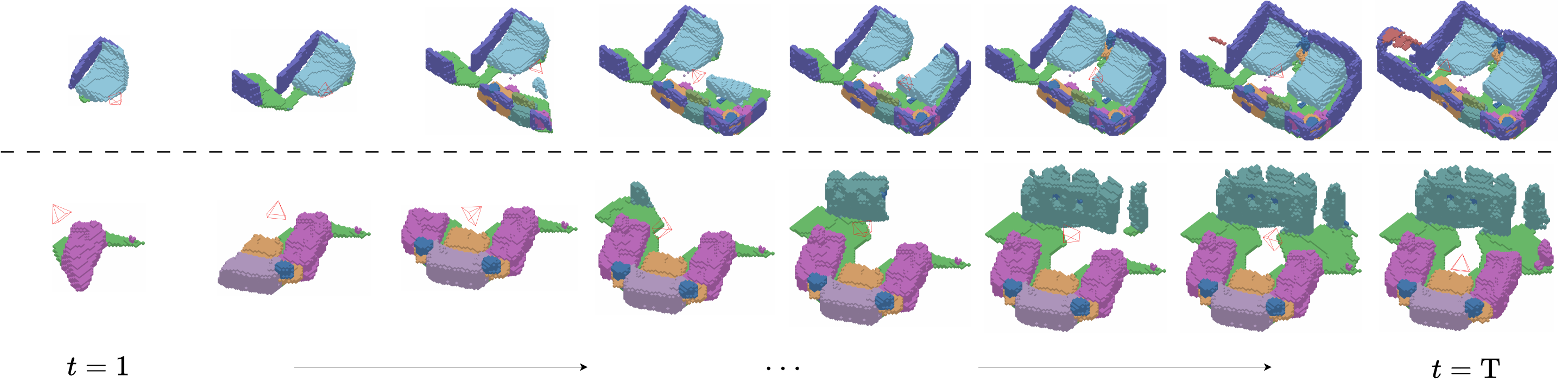}
      \captionsetup{font={small}}
      \caption{\textbf{Qualitative results of embodied occupancy prediction on EmbodiedOcc-ScanNet.} Our {\methodname} successfully achieves high-quality online occupancy prediction under diverse exploration trajectories.}
      \label{fig:vis_embodied_pred}
      \vspace{-10pt}
\end{figure*}

\subsection{Datasets and Metrics}
We evaluate our {\methodname} on two challenging benchmarks: the Occ-ScanNet~\cite{yu2024monocular} dataset for local occupancy prediction and the EmbodiedOcc-ScanNet~\cite{wu2025embodiedocc} dataset for embodied occupancy prediction. The Occ-ScanNet dataset provides voxelized observations in a 60 × 60 × 36 grid (covering a 4.8 m × 4.8 m × 2.88 m region in front of the camera), with annotations over 12 semantic classes. Built upon Occ-ScanNet, the EmbodiedOcc-ScanNet dataset further offers dense semantic labels for entire scenes, enabling evaluation in embodied settings. We adopt commonly used metrics for occupancy prediction, including mIoU and IoU to assess accuracy, as well as memory usage and inference time to evaluate efficiency.

\subsection{Implementation Details}
We train {\methodname} on 4 RTX 3090 GPUs with a total batch size of 16. Following the linear scaling rule, the learning rate is set to 4e-4, i.e., twice that of EmbodiedOcc~\cite{wu2025embodiedocc}. For local occupancy prediction, we follow prior works and train for 10 epochs on Occ-ScanNet. For embodied occupancy prediction, we freeze the local prediction modules and optimize the TLA module for only 1 epoch on EmbodiedOcc-ScanNet, much more efficient than previous settings.

\subsection{Results on Public Benchmarks}

\PAR{Local Occupancy Prediction.} 
Table~\ref{table:local_pred_main} compares local occupancy prediction on Occ-ScanNet. VEOcc achieves $64.55$ IoU and $55.49$ mIoU, outperforming GPOcc-DPT by $+7.59/+3.61$ and TGSFormer by $+0.13/+0.76$ IoU/mIoU, respectively. GPOcc-VGGT~\cite{Zhou_2026_CVPR} reports $63.14/56.19$ with $942.31$M parameters, over $5.3\times$ our $177.5$M (Table~\ref{tab:robot_latency}), and relies on a large pretrained VGGT~\cite{Wang_2025_CVPR} backbone. Given this different scaling and pretraining regime, we report it separately as a large-foundation-model reference rather than a directly comparable baseline. These results highlight the advantage of voxel-centric representations in preserving geometric structures, semantic boundaries, and fine-grained scene details. Overall, VEOcc achieves state-of-the-art performance among models of comparable scale.

\PAR{Embodied Occupancy Prediction.}
Table~\ref{table:embodied_pred_main} compares embodied occupancy prediction on EmbodiedOcc-ScanNet. VEOcc achieves $62.21$ IoU and $53.00$ mIoU, outperforming GPOcc-DPT and TGSFormer by $+5.82/+1.78$ and $+7.79/+7.71$ IoU/mIoU, respectively, and achieving state-of-the-art embodied performance among models of comparable scale. For the same model-scale consideration discussed above, GPOcc-VGGT~\cite{Zhou_2026_CVPR} ($61.41/55.39$) is not included in the main comparison. Compared with local prediction, the embodied setting further requires continuous integration of sequential observations into a coherent global representation. The strong absolute performance shows that VEOcc retains high prediction accuracy under this more challenging setting.

\begin{figure*}[t]
      \centering
      \includegraphics[width=\linewidth]{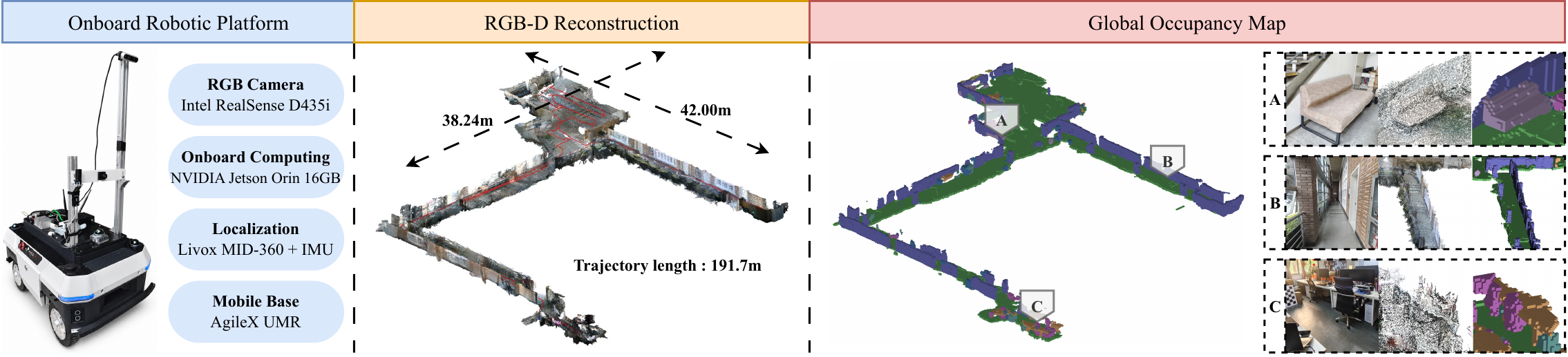}
      \captionsetup{font={small}}
      \caption{\textbf{Onboard robotic deployment and long-trajectory mapping.}
        Left: the robotic platform, where {\methodname} uses monocular RGB images for occupancy prediction and LiDAR--inertial odometry only for localization. Middle: the RGB-D reconstruction and trajectory of a $191.7\,\mathrm{m}$ exploration sequence spanning approximately $42.00\,\mathrm{m}\times38.24\,\mathrm{m}$. Right: the incrementally constructed global semantic occupancy map, with regions A--C enlarged to compare the RGB observations, geometric reconstructions, and predicted occupancy.}
      \label{fig:real_world}
      \vspace{-10pt}
\end{figure*}
\begin{figure*}[t]
      \centering
      \includegraphics[width=0.95\linewidth]{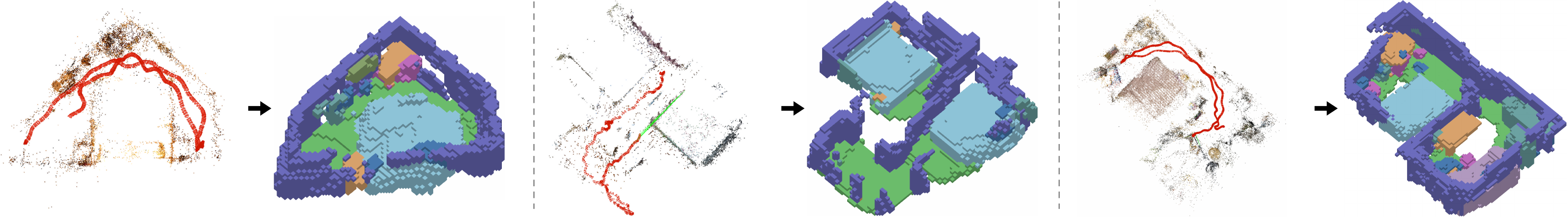}
      \captionsetup{font={small}}
      \caption{\textbf{Zero-shot real-world generalization on self-collected indoor sequences.} We show the COLMAP sparse reconstruction and our incrementally generated global semantic occupancy map for a single‑room layout (left) and double‑room layouts (middle, right). Without fine-tuning or scene priors, {\methodname} produces coherent global occupancy maps across unseen environments under rich viewpoint variations.}
      \label{fig:zero_shot}
      \vspace{-10pt}
\end{figure*}

\PAR{Qualitative Results.}
Fig.~\ref{fig:vis_local_pred} presents qualitative comparisons of local occupancy prediction results. Compared with prior methods, {\methodname} achieves superior overall performance, particularly in challenging regions such as boundary structures, distant objects, and occluded areas highlighted by the red bounding boxes. Building upon accurate local prediction, Fig.~\ref{fig:vis_embodied_pred} further demonstrates that VEOcc maintains robust embodied occupancy prediction under diverse trajectories and camera viewpoints. Moreover, Fig.~\ref{fig:GFS} validates the effectiveness of the proposed online update strategy, which progressively corrects unreliable observations and improves global occupancy consistency over time.

\subsection{Real-World Deployment and Zero-Shot Generalization.}
\PAR{Onboard Robotic Deployment.}
As shown in Fig.~\ref{fig:real_world}, we deploy {\methodname} on a wheeled robot for online semantic occupancy mapping, with all processing performed onboard. Occupancy prediction takes only monocular RGB images as sensor input, without using the camera's depth stream, while LiDAR--inertial odometry supplies metric camera poses. At $1\,\mathrm{Hz}$, the end-to-end pipeline averages $783.75\,\mathrm{ms}$ per frame (Table~\ref{tab:robot_latency}), supporting online operation. Along a $191.7\,\mathrm{m}$ trajectory, the system incrementally constructs a sparse global map spanning $42.00\times38.24\times2.08\,\mathrm{m}$ and containing $861{,}782$ voxel states. The final map occupies $118.348\,\mathrm{MiB}$ of logical memory (approximately $144\,\mathrm{B}$ per voxel), with a total process RSS of $5680.3\,\mathrm{MiB}$, demonstrating practical onboard deployment and long-horizon scalability.
\begin{table}[t]
\centering
\captionsetup{font=small, skip=4pt}
\small
\setlength{\tabcolsep}{3pt}

\caption{\textbf{Cross-platform runtime} of {\methodname}.
Latency is reported as mean per-frame inference time in milliseconds.}
\label{tab:robot_latency}

\color{black}
\begin{tabular}{lcccc}
\toprule
& & \multicolumn{2}{c}{RTX 3090}
& \multicolumn{1}{c}{Jetson Orin 16GB} \\
\cmidrule(lr){3-4}
\cmidrule(lr){5-5}
Method & $\text{N}_{\text{param}}$ & Local & Embodied & End-to-end \\
\midrule
{\methodname} & 177.5M & 169 ms & 175 ms & 783.75 ms \\
\bottomrule
\end{tabular}
\end{table}



\begin{table}[t]
\centering
\captionsetup{font={small}, skip=6pt}
\small
\caption{Comparison between different state update strategies.}
{\small

\color{black}
\begin{tabular*}{\linewidth}{l@{\hspace{1em}} |l@{\hspace{2.8em}} |@{\hspace{1em}} c@{\hspace{1em}} c}
\toprule
 & Fusion Strategy & IoU & mIoU \\
\midrule
(a) & Overwrite & 58.38 & 48.31 \\
(b) & Highest Probability & 59.33 & 50.87 \\
(c) & Weighted Average on Logit & 61.90 & 52.81 \\
(d) & Weighted Average on Probability & \textbf{62.21} & \textbf{53.00} \\
\bottomrule
\end{tabular*}
}
\vspace{-10pt}
\label{tab:fusion_strategy}
\end{table}

\PAR{Zero-Shot Handheld Generalization.}
We further apply {\methodname} zero-shot to self-collected handheld smartphone sequences from multiple unseen single- and double-room environments. Handheld capture introduces richer viewpoint and camera-motion variations than the robot-mounted setting, providing a complementary and more diverse test of real-world generalization. Camera trajectories are estimated with COLMAP~\cite{schonberger2016structure}, and the model trained solely on EmbodiedOcc-ScanNet is used directly without fine-tuning. \rev{For each handheld sequence, the metric scale is recovered from a measured scene distance and the corresponding COLMAP reconstruction distance.}  As shown in Fig.~\ref{fig:zero_shot}, {\methodname} consistently reconstructs coherent global semantic occupancy maps across varied layouts and viewpoints, demonstrating strong generalization beyond the training domain.
























\begin{table}[t]
\centering
\captionsetup{font={small}, skip=6pt}
\caption{Ablation on embodied prediction modules.}
\small

\color{black}
\begin{tabular*}{\linewidth}{ l@{\hspace{0.9em}} |@{\hspace{0.9em}}c@{\hspace{0.9em}}c@{\hspace{0.9em}} |@{\hspace{0.9em}}c@{\hspace{0.9em}}c@{\hspace{0.9em}} |@{\hspace{0.9em}} c@{\hspace{0.9em}}c}

\toprule
 & TLA & $\text{RCM}$ & ${\alpha}_\text{RCM}$ & ${\beta}_\text{RCM}$ & IoU & mIoU \\
\midrule
(a) & \xmark & \xmark & -    & -    & 60.48 & 51.93 \\
(b) & \cmark & \xmark & -    & -    & 61.23 & 52.27 \\
(c) & \cmark & \cmark & 0.1 & 0    & 62.03 & 52.87 \\
(d) & \cmark & \cmark & 0    & 1.5  & 61.43 & 52.42 \\
(e) & \cmark & \cmark & 0.1  & 1.5  & \textbf{62.21} & 53.00 \\
\midrule
(f) & \cmark & \cmark & learnable  & 1.5 & 62.00 & 52.89 \\
(g) & \cmark & \cmark & 0.1    & learnable  & 62.20 & \textbf{53.02} \\
(h) & \cmark & \cmark & learnable    & learnable  & 62.02 & 52.91 \\
\bottomrule

\end{tabular*}
\label{tab:online_modules}
\vspace{-10pt}
\end{table}

\subsection{Ablation Studies}
We conduct comprehensive ablation studies on the EmbodiedOcc-ScanNet dataset.

\PAR{Ablation on embodied prediction modules.}
Table~\ref{tab:online_modules} evaluates TLA and RCM under the same local predictor. Adding TLA improves IoU/mIoU from 60.48/51.93 to 61.23/52.27, confirming the benefit of short-term temporal aggregation. Incorporating RCM further improves performance, where depth-aware decay ($\alpha_{\mathrm{RCM}}$) contributes more strongly than boundary-aware decay ($\beta_{\mathrm{RCM}}$), and combining both achieves 62.21 IoU and 53.00 mIoU. We further examine learnable decay parameters and observe only marginal variations compared with the fixed setting. This indicates that the performance gain of RCM mainly comes from its reliability-aware modulation mechanism rather than parameter tuning.

\PAR{Analysis of incremental state update strategies.}
Table~\ref{tab:fusion_strategy} evaluates different incremental state update strategies for aggregating temporal local predictions in the CSU module. Naive strategies relying on single observations such as overwrite (a) and selecting the highest probability (b) fail under multi-view inconsistencies. In contrast, aggregation-based methods (c–d) significantly improve performance, with weighted averaging consistently outperforming simpler alternatives. Notably, performing weighted averaging on probabilities (d) achieves the best results, consistently surpassing logit-based fusion (c). This confirms our theoretical design: while short-term fusion operates safely on logits, aggregating probabilities effectively normalizes logit variance across distant temporal observations for stable long-term integration.

\section{Conclusion}
In this work, we presented {\methodname}, a voxel-centric framework that combines frame-wise voxel prediction with a Spatio-Temporal-Aware Online Update Strategy comprising TLA, RCM, and CSU, enabling direct maintenance and open-ended expansion of a sparse global semantic voxel map. Extensive experiments demonstrate state-of-the-art performance among models of comparable scale on both local and embodied benchmarks, while real-world deployment further validates practical online operation and generalization across diverse unseen environments. VEOcc remains susceptible to pose drift and scene dynamics, and early mapping errors may persist. Future work will develop robust fusion strategies capable of revising accumulated global states.






{
\AtNextBibliography{\scriptsize}
\printbibliography

}

\end{document}